\documentclass[conference]{IEEEtran}
\usepackage{cite}

\usepackage{url}

\ifCLASSINFOpdf
   \usepackage[pdftex]{graphicx}
\else
\fi
\usepackage{amsmath}

\newcommand{\vh}[0]{\mathbf{h}}
\newcommand{\vx}[0]{\mathbf{x}}

\usepackage{caption}
\usepackage{subcaption}
\usepackage{multirow}

\usepackage{amssymb}
\IEEEoverridecommandlockouts 

\begin{document}
%
\title{
DeepTrend: A Deep Hierarchical Neural Network for Traffic Flow Prediction
}

\author{Xingyuan Dai$^{1,2}$, Rui Fu$^{3}$, Yilun Lin$^{1,2}$, \\
	Li Li$^{3}$, \textit{Fellow, IEEE}, and Fei-Yue Wang$^{1}$, \textit{Fellow, IEEE}
	\thanks{$^{1}$The State Key Laboratory of Management and Control for Complex Systems, Institute of Automation, Chinese Academy of Sciences, Beijing, 100190, China}%
	\thanks{$^{2}$University of Chinese Academy of Sciences, Beijing, 100049, China}%
	\thanks{$^{3}$Department of Automation, Tsinghua University, Beijing, 100084, China}
	\thanks{**Li Li is the corresponding author of this paper. {\tt\small E-mail: li-li@tsinghua.edu.cn}}
}

\maketitle

\begin{abstract}
In this paper, we consider the temporal pattern in traffic flow time series, and implement a deep learning model for traffic flow prediction. Detrending based methods decompose original flow series into trend and residual series, in which trend describes the fixed temporal pattern in traffic flow and residual series is used for prediction. Inspired by the detrending method, we propose DeepTrend, a deep hierarchical neural network used for traffic flow prediction which considers and extracts the time-variant trend. DeepTrend has two stacked layers: extraction layer and prediction layer. Extraction layer, a fully connected layer, is used to extract the time-variant trend in traffic flow by feeding the original flow series concatenated with corresponding simple average trend series.  Prediction layer, an LSTM layer, is used to make flow prediction by feeding the obtained trend from the output of extraction layer and calculated residual series. To make the model more effective, DeepTrend needs first pre-trained layer-by-layer and then fine-tuned in the entire network. Experiments show that DeepTrend can noticeably boost the prediction performance compared with some traditional prediction models and LSTM with detrending based methods. 

\end{abstract}

\IEEEpeerreviewmaketitle

\section{Introduction}

Traffic flow prediction is one of the major tasks of intelligent transportation systems (ITSs) that should be resolved \cite{Wang2010}. It is strongly needed for individuals, companies, governments and so on to make decisions in time according to different conditions of traffic flow. However, accurate and real-time traffic prediction remains challenging and unsolved for many decades due to its stochastic and nonlinear feature. Traditional methods mainly use linear models like autoregressive integrated moving average (ARIMA)  \cite{Ahmed1979b, Levin1980, Lee1999, Kamarianakis2003, Williams2003} and multi-variable linear regression (MVLR) \cite{Li2015, Chrobok2004}, and some machine learning models like support vector regression (SVR) \cite{Jin2007a} to predict incoming traffic flow but cannot consider the entire features in traffic flow and perform not very well.

In recent years, some deep learning methods for traffic flow prediction are put forward like stacked autoencoders (SAEs)  \cite{Lv2014}, long short-term memory network (LSTM) \cite{Tian2015}, deep belief network (DBN)  \cite{Huang2014}, etc., and have good performance. On one hand, these models generally have complex network structure which can fit nonlinear parts in traffic flow series. On the other hand, some deep learning models like LSTM and gated recurrent unit (GRU) \cite{Fu2016} are designed especially for the time series which are adept in dealing with traffic flow. 

In this paper, we explore whether deep networks like LSTM can learn the temporal patterns existed in flow time series, which is of great importance for traffic prediction. However, the experiments show that LSTM has similar prediction performance with some traditional machine learning models. To make LSTM more effective in flow prediction, we introduce detrending based methods, which are frequently used in traffic flow prediction nowadays \cite{Chen2012,Li2015,Li2015a,Li2014}. It is based on the hypothesis that there exists a certain temporal pattern trend in traffic flow time series and can be separated from the remaining fluctuations. 
 So researchers often assume there exists invariant periodic trend in traffic flow time series. Some methods were used to retrieve intra-day or seasonal trend via simple-average, principal component analysis (PCA) or wavelet methods. 

 Inspired by the idea of detrending, we propose a well-designed deep network architecture named DeepTrend. DeepTrend has two kinds of hidden layers: extraction layer and prediction layer. The extraction layer is used to learn to extract the time-variant trend, and the prediction layer is used to predict the incoming flow by feeding the extracted trend and calculated residual series. Experiments show that DeepTrend outperforms other baseline based on original flow data or detrending methods.

The rest of this paper is organized as follows. Section II reviews the studies on short-term traffic flow prediction. Section III proposes the DeepTrend architecture for traffic flow prediction and the detrending based method. Section IV discusses the experiment design and performance of the proposed architecture, and comparison with several selected models. Finally,  Section V concludes the paper.

\section{Literature Review}
In general, traffic flow prediction approaches can be divided into two major categories: parametric approach and nonparametric approach. 

The main parametric approach includes ARIMA \cite{Ahmed1979b} model, 
MVLR \cite{Li2015, Chrobok2004}.
The model architecture of these approaches is predetermined based on the certain theoretical assumptions and the model parameters should be calculated by empirical data. ARIMA model is based on the assumption that the traffic condition is in a stationary process. It was first used for short-term traffic flow prediction in the 1970s \cite{Ahmed1979b}, and then ARIMA (0, 1, 1) \cite{Levin1980} was found the most statistically significant for flow prediction. Moreover, some improved ARIMA models like subset ARIMA \cite{Lee1999}, space-time ARIMA \cite{Kamarianakis2003} and seasonal ARIMA (SARIMA) \cite{Williams2003} were also proposed to forecast traffic flow.
The parametric approach has simple and explicit architecture and takes a little time to obtain the results. 

However, due to the stochastic and nonlinear feature in traffic flow, the parametric approach with linearity cannot present a high performance for traffic flow prediction. Therefore, researchers have paid much attention to the nonparametric approach such as $k$-NN \cite{Davis1991}, SVR \cite{Jin2007a}, online support vector regression (OL-SVR) \cite{Castro-Neto2009}, random forests regression (RF) \cite{Leshem2007a}, gradient boosting regression\cite{Friedman2001}. A variety of artificial neural network (ANN) models were proposed to predict traffic flow and perform well \cite{Vlahogianni2005,Chan2012,Zhong2005}. Recently, with the development of deep learning, many deep learning models were applied to traffic flow prediction.  SAE \cite{Lv2014}, DBN \cite{Huang2014}, LSTM \cite{Tian2015} and GRU \cite{Fu2016}  model were proposed in traffic flow forecasting and got superior performance. However, these recent studies do not further explore to extract the intra-patterns of flow series in models, which needs to be concerned for better traffic flow prediction.

\section{Methodology}

\subsection{Recurrent Neural Network (RNN)}
\begin{figure}[!htb]
	\centering
	\includegraphics[width=3.4in]{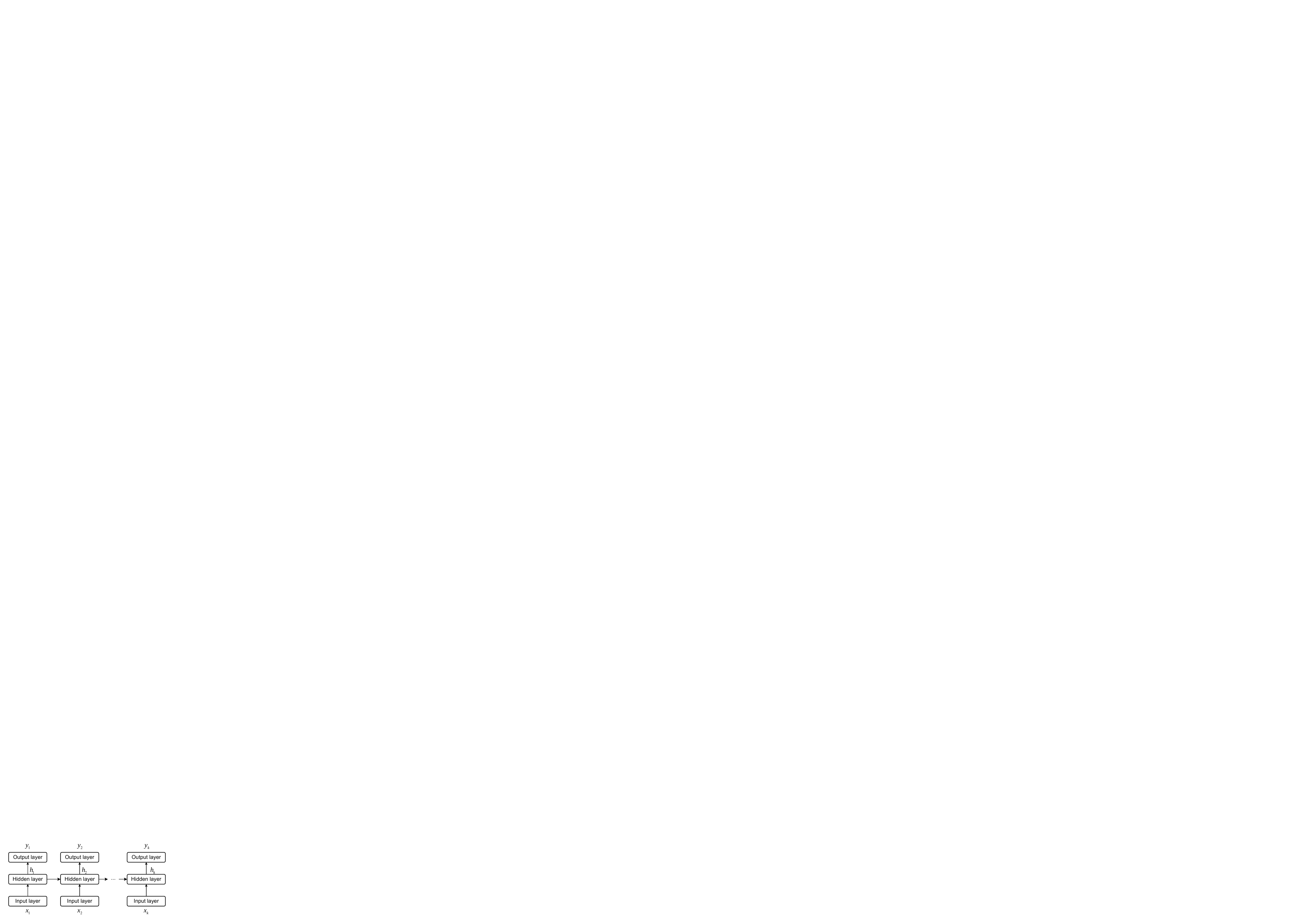}
	\caption{The structure of RNN.}
	\label{fig: RNN}
\end{figure}

The RNN \cite{Goodfellow-et-al-2016} is a generation of the feedforward neural networks which is adept in dealing with sequences. The structure of RNN is shown in Fig. \ref{fig: RNN}. Given a general input sequence $(x_1, x_2, ..., x_k)$ where $x_i \in \mathbb{R}^d$, a hidden state is obtained at each time step, resulting in a hidden sequence $(h_1, h_2, ..., h_k)$. The hidden state at time step $t$ is calculated by the function 
\begin{align}
\label{eq: RNN}
h_t=f\left(x_t, h_{t-1}\right)
\end{align}
in which $x_t$ is the current input and $h_{t-1}$ is the previous hidden state. Then the optional output at each time step is calculated by $y_t=g(h_t)$. The output of RNN can be a sequence as $(y_1, y_2, ..., y_k)$ or a single value $y_k$ which is dependent on the objective of the problems.

The simple RNN calculates the output at each time step, making the network very deep. It is hard for them to train and capture the dependence of the input sequence. Thus, the structure of hidden layer is essential for them. 

\subsection{Long Short-Term Memory network (LSTM)}

\begin{figure}[!htb]
	\centering
	\includegraphics[width=2in]{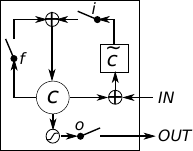}
	\caption{Long Short-Term Memory}
	\label{fig: LSTM_units}
\end{figure}

LSTM \cite{Hochreiter1997} is a special kind of RNN,  designed to learn long-term dependencies. It has a complex structure named LSTM unit in its hidden layer which contains three gates namely input gate, forget gate and output gate to protect and control the unit state. The LSTM unit is shown in Fig \ref{fig: LSTM_units}, in which IN represents the input data and the previous unit's output.

Denote that the input is $x_t$ and the hidden units output is $h_t$ at time step $t$ and their previous output is $h_{t-1}$. For the $j$-th LSTM unit, the input gate $i_t^j$,  forget gate $f_t^j$ and output gate $o_t^j$ can be calculated using the following equations:

\begin{align}
\label{eq: lstm_memory_up}
	i_t^j & = \sigma\left( W_{xi} x_t + W_{hi} h_{t-1} + {b_i} \right)^j\\
	f_t^j & = \sigma\left( W_{xf} x_t + W_{hf} h_{t-1} + {b_f} \right)^j \\
	o_t^j & = \sigma\left( W_{xo} x_t + W_{ho} h_{t-1} + {b_o} \right)^j 
\end{align}

where $\sigma$ is a logistic sigmoid function, $W$ terms are weight matrices, and $b$ terms are bias vectors.

Unlike traditional recurrent unit, each $j$-th LSTM unit maintains a memory  $c_t^j$ at time $t$. The memory cell  $c_t^j$ is updated by
\begin{align}
\label{eq: lstm_memory_up2}
c_t^j = f_t^j c_{t-1}^j + i_t^j \tilde{c}_t^j
\end{align}
where new memory content is 
\begin{align}
	\tilde{c}_t^j & = \tanh\left( W_{xc} x_t + W_{hc} h_{t-1} + {b_c} \right)^j
\end{align}

The LSTM unit output is computed by
\begin{align}
h_t^j & = o_t^j \tanh \left( c_t^j \right) 
\end{align}

\subsection{LSTM Network for Traffic Flow Prediction}

\begin{figure}[!htb]
	\centering
	\includegraphics[width=3.5in]{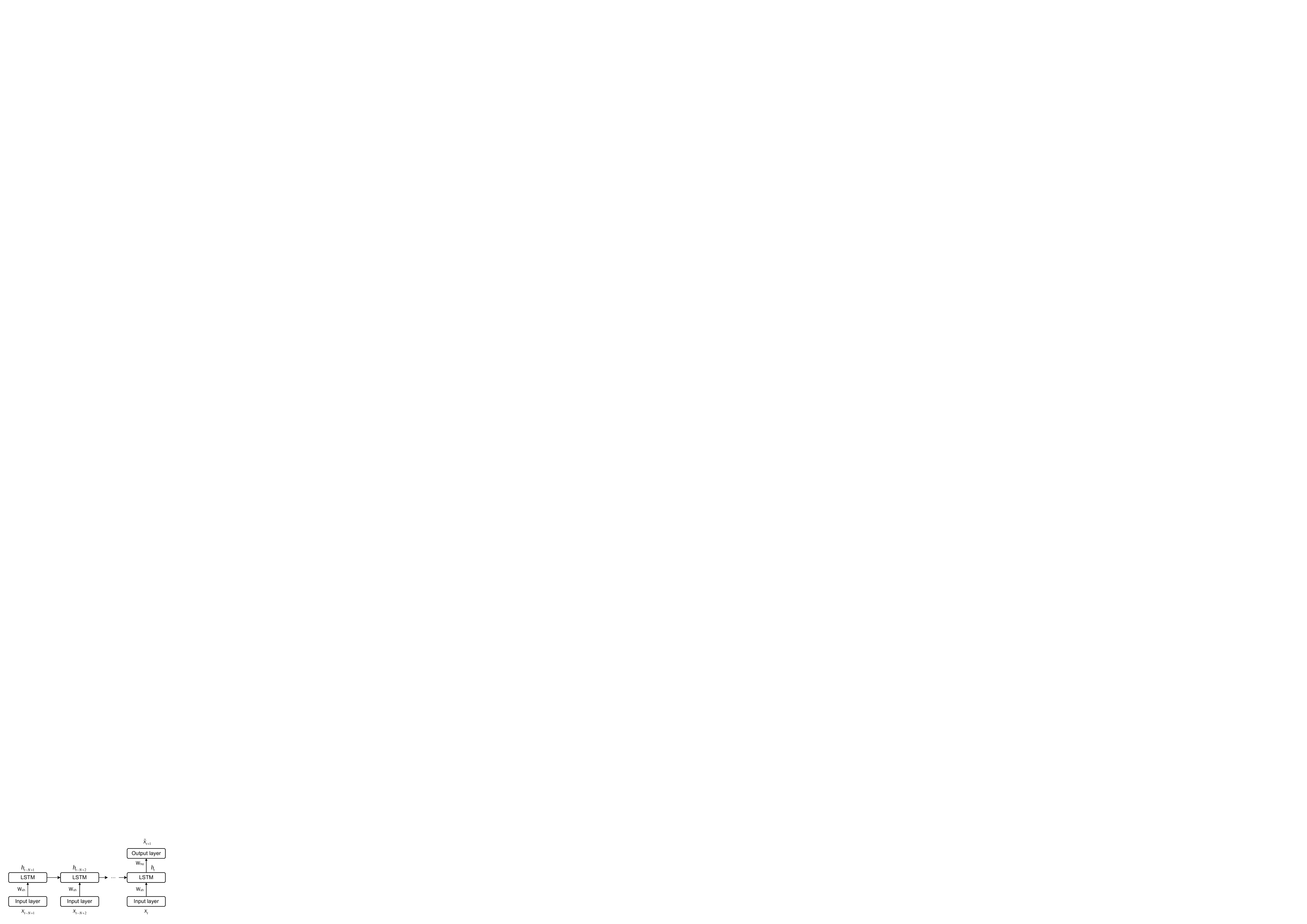}
	\caption{The structure of one-layer LSTM network for traffic prediction.}
	\label{fig: LSTM}
\end{figure}

We apply one-layer LSTM to traffic flow prediction. The main architecture is shown in Fig \ref{fig: LSTM}. At time $t$, the input of the network is the observed historical traffic data which we use  the previous $N$ steps data as $\vx=(x_{t-N+1}, x_{t-N+2}, ..., x_t)$ and the output  $\hat{x}_{t+1}$ is the predicted traffic flow in next time step. We can get the hidden unit output $h_t$ using the above equations, and the output of the network can be calculated as
\begin{align}
& \hat{x}_{t+1}=W_{ho} \vh_t+b
\end{align}
where $W_{ho}$ is the weight matrix between the hidden layer and output layer and $b$ is bias term. Then, we use Back Propagation Through Time (BPTT) \cite{Werbos1990} algorithm to train our model.

\subsection{Detrending Based Prediction}

Detrending \cite{Chen2012,Li2015,Li2015a,Li2014} is widely used in analyzing and predicting traffic flow series. The goal of detrending is to remove the periodic trend that may influence traffic prediction and using the residual time series to make predictions. In our work, we make prediction without distinguishing between weekday and weekend. Therefore, the daily periodic trend can not be considered because there are huge different patterns of traffic flow in weekday and weekend, and we determine the weekly-periodic trend.

The easiest way to calculate the trend is to use the average of periodic traffic flow time series collected in the same station, which is called simple average trend.

Let $y^i_{j-k}$ denote the $k$th sample point data at station $i$ in $j$th week. The traffic time series in $N$ continuous weeks can be written as a series of one-dimensional vectors 

 \begin{align}
 \label{eq: Detrending}
& Y_{1}^i=\left[ y_{1-1}^i, y_{1-2}^i, ..., y_{1-n}^i\right], ..., Y_{N}^i=\left[ y_{N-1}^i, y_{N-2}^i, ..., y_{N-n}^i \right]
 \end{align}
where $n$ is the number of sample data points per week. If the sample time interval is 5 minutes, we have $n=288\times7$.

The simple average trend over past $D$ weeks can be calculated as
 \begin{align}
 & Y_{Average}^i=\left[ \frac{1}{D} \sum_{j=N-D+1}^{N} y_{j-1}^i,  ..., \frac{1}{D} \sum_{j=N-D+1}^{N} y_{j-n}^i \right]
\end{align}
where $D=N$ indicates the average for all sample weeks.

Then, we can obtain the residual time series $R^i_j=\left[ R_{j-1}^i, R_{j-2}^i, ..., R_{j-n}^i \right]$ by subtracting the simple average trend from the original time series as
 \begin{align}
R^i_j = Y^i_{j}- Y_{Average}^i
\end{align}

The residual time series instead of original ones are finally fed into the prediction models in detrending based methods.

\renewcommand\arraystretch{2}
\begin{table*}[!htb]
	\caption{Performance comparison of different models.}
	\label{tab: model_compare}
	\centering
	\begin{tabular}{c||c|c|c|c|c||c|c|c|c|c||c}
		\hline
		\hline
		& ARIMA-O  & MVLR-O  & SVR-O   & RF-O    & LSTM-O  & ARIMA-D  & MVLR-D  & SVR-D   & RF-D    & LSTM-D  & DeepTrend \\
		\hline
		MSE & 1129.89 & 1138.60 & 1062.94 & 1110.54 & 1072.23 & 1028.64 & 1036.78 & 1031.51 & 1085.31 & 1024.43 & \textbf{984.47}    \\
		MAE & 22.85   & 23.12   & 22.18   & 22.64   & 22.44   & 21.54   & 21.62   & 21.44   & 22.03   & 21.49   & \textbf{21.21}     \\
		\hline
		\hline
	\end{tabular}
\end{table*}

\subsection{DeepTrend}
\begin{figure}[!htb]
	\centering
	\includegraphics[width=3.5in]{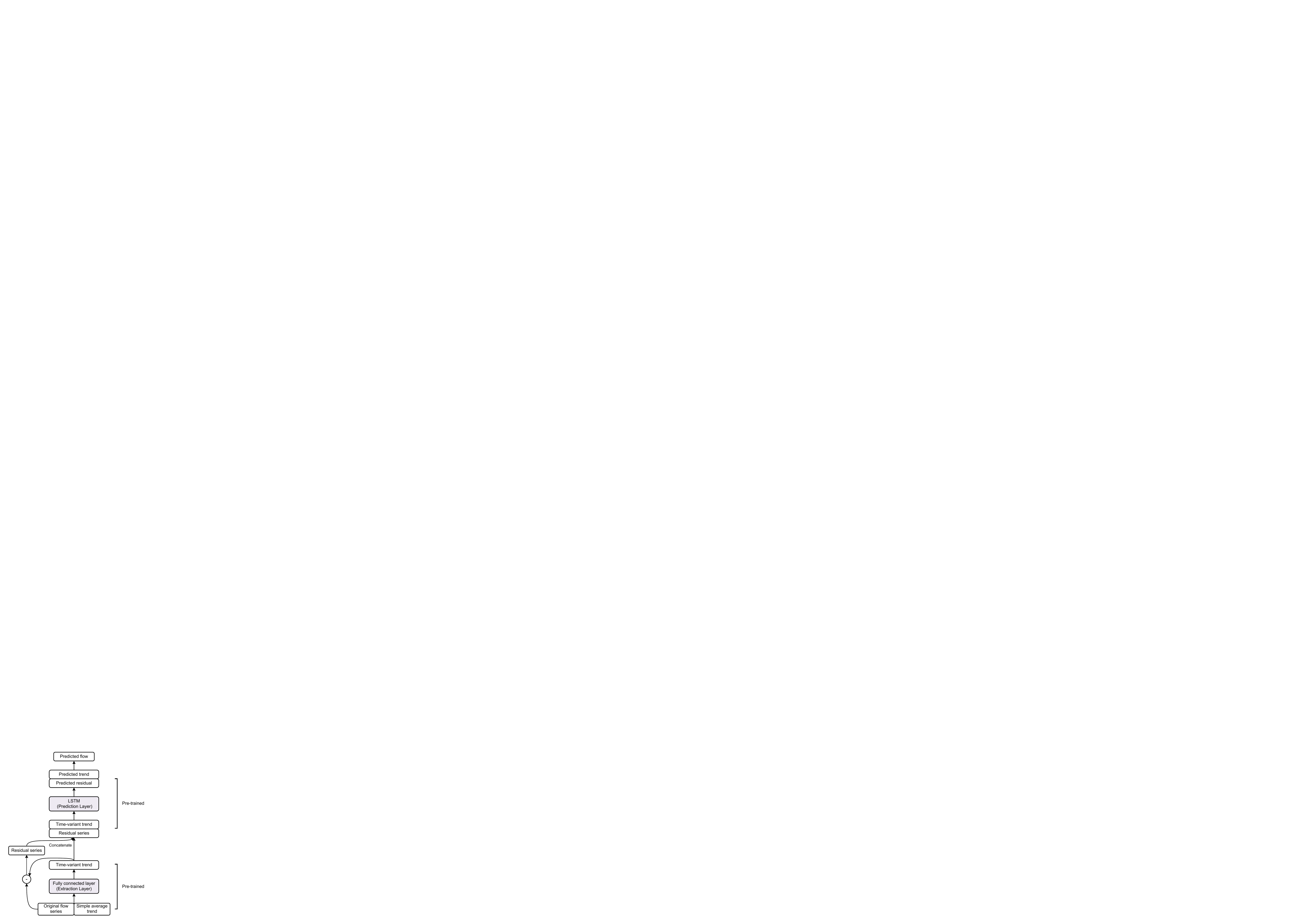}
	\caption{The structure of DeepTrend.}
	\label{fig: Deeptrend}
\end{figure}
Simple average trend is periodic and fixed. Even there are big difference between the traffic flows at the same time in different weeks, their corresponding simple average trend is still same. This will make the residual series hard to forecast if there are large offsets between original flow and trend. 

To solve the problem and make it better for the predictor to learn the temporal pattern existed in flow time series, we propose DeepTrend, which is used to better capture the time-variant trend and lift the prediction performance. 

As shown in Fig. \ref{fig: Deeptrend},  DeepTrend contains two kinds of hidden layers: extraction layer and prediction layer, in which the extraction layer is a fully connected layer and the prediction layer is an LSTM layer. 

Extraction layer is designed to extract the time-variant trend by feeding the original flow series and corresponding simple average trend series. Prediction layer is used to predict the incoming traffic flow which is an addictive combination of the predicted trend and residual. The prediction layer is fed by the obtained time-variant trend series and the residual series calculated by subtracting the obtained trend from original flow. In a sense, the DeepTrend can be regarded as a special detrending method which decomposes the flow time series into trend and residual series.

In order to make the network learn the flow patterns better, avoid the deep network falling into a local minimum during training, and speed up the convergence, we implement the method that first pre-training network layer-by-layer and then fine-tuning the entire network. That is to say, we first train the extraction layer by feeding the original flow series concatenated with simple average trend series as inputs and simple average trend series as output to make the network first reconstruct the simple average trend and also carry the information from original flow series. Then we use the obtained trend from output of extraction layer and calculated residual series as two features of inputs and predicted trend and residual values as outputs to train prediction layer. Finally, we train the total network using a small learning rate by feeding the original flow series and simple average trend series as input and the predicted flow value in next time step as output. 

The pre-training process makes the extraction layer and prediction layer adapt to their task quickly, and fine-tuning process will further decrease the training loss and make the network finally converge. This training scheme enables DeepTrend to better extract the time-variant trend, and further improve the performance for traffic flow forecasting.

\section{Experiments}
\subsection{Data Sources and Data-Pre-Processing}
The experimental data of traffic flow are obtained from Caltrans Performance Measurements Systems (PeMS) \cite{pems}, which are collected from 3941 stations every 5 min in district 4 of freeway systems across California. Our model is applied to the data in the first 16 weeks of 2016. The first 12 week' data are selected as the training set, and the remaining 4 week' data are selected as the test set.

Because there are some missing traffic flow data in some stations, we select 1397 stations from the original data set  whose missing flow data in 2016 are less than $1\%$ to make analysis, and impute the missing traffic flow data of these stations using simple average trend. 

Considering the limited computational resource, in our experiments, we select 50 stations to forecast traffic flow. Before feeding into the model, the flow data in each station are first normalized to be 0 mean and 1 standard deviation.

\subsection{Performance Indexes}
To evaluate the performance of the proposed model, we use two performance indexes, which are mean square error (MSE) and  mean absolute error (MAE). These indexes are defined as
\begin{align}
&\text{MSE}=\frac{1}{T} \sum_{t=1}^{T}\left(y_t-\hat{y}_t\right)^{2} \\
&\text{MAE}=\frac{1}{T}\sum_{t=1}^{T} \left| {y_t - \hat{y}_t}\right| 
\end{align}
where $y_t$ and $\hat{y}_t$ are the actual traffic flow and predicted traffic flow.

\subsection{Predictor Architecture Settings}

Considering the temporal correlation, we use the previous $N$ steps data as $\vx=(x_{t-N+1}, x_{t-N+2}, ..., x_t)$ to predict traffic flow in next time step denoted as $\hat{x}_{t+1}$.  In our experiment, $N$ is set to 12. That is to say, we use the history values within the last 1 hour to predict traffic flow in next 5 minutes.
 
There are several parameters in our prediction architecture that need defining and tuning.
For DeepTrend, the extraction layer contains 128 neurons units and the prediction layer contains 128 LSTM units. The activation functions of the two layers are both ReLU. The optimization algorithm is using Adam \cite{kingma2014adam}. The learning rates in pre-training the extraction layer and prediction layer are set to 0.001 and 0.005, respectively. Their numbers of pre-training epochs are 20 and 10, respectively. In fine-tuning, the learning rate is set to 0.00002 and the number of fine-tuning epochs is 7.  
For ARIMA model, we use ARIMA(12, 0, 1) as the comparative model. 
For SVR, the penalty parameter $C$ is 1.0, $\epsilon $ is 0.1 and RBF kernel is used. 
For random forests (RF), the number of trees in the forest and maximum depth are both 10.
For LSTM, we adopt one-layer LSTM network which has 128 LSTM units. The activation function for the hidden layer is ReLU, and for the output layer is a linear function. It also uses Adam to train the model. The learning rate is set to 0.001 and the number of training epochs is 20. 

It should be pointed out that there are some other parameters had been tested and the above settings are among the best ones.

We make use of Keras \cite{chollet2015keras} library with Tensorflow \cite{abadi2016tensorflow} as backend to implement DeepTrend and LSTM, and use scikit-learn \cite{pedregosa2011scikit} library to implement MVLR, SVR, and RF.

\subsection{Experimental Results}

We evaluate the prediction performance of the proposed DeepTrend with the traditional models like ARIMA, MVLR, SVR, RF, and deep network LSTM. The comparative models are based on original flow and detrending methods. 

Table \ref{tab: model_compare} shows the results of different models for a 5-min flow prediction, in which Model-O denotes that the model is using original data and Model-D denotes detrending based model.

The results show that (1) simply using LSTM still cannot significantly boost the prediction performance; (2) detrending based models significantly outperform the original data based models, and (3) the proposed DeepTrend perform better than detrending based models.

If the original flow time series data are used in prediction, SVR performs best in terms of MSE and MAE. Although LSTM as a deep network is adept in dealing with time series and learning the data representation, in the experiment, simply using an LSTM network still has not learned most intra-pattern of original flow series and is not dominant compared with the traditional model SVR. 

If detrending based methods are used, all models have gained significant boosting in prediction performance. This indicates that trend served as a key component in temporal patterns of flow plays an import role in the traffic flow prediction. For detrending based models, LSTM performs best in MSE term and SVR performs best in MAE term, and the differences of two indexes between them are not very large.

\begin{figure}[!htb]
	\centering
	\includegraphics[width=3.5in]{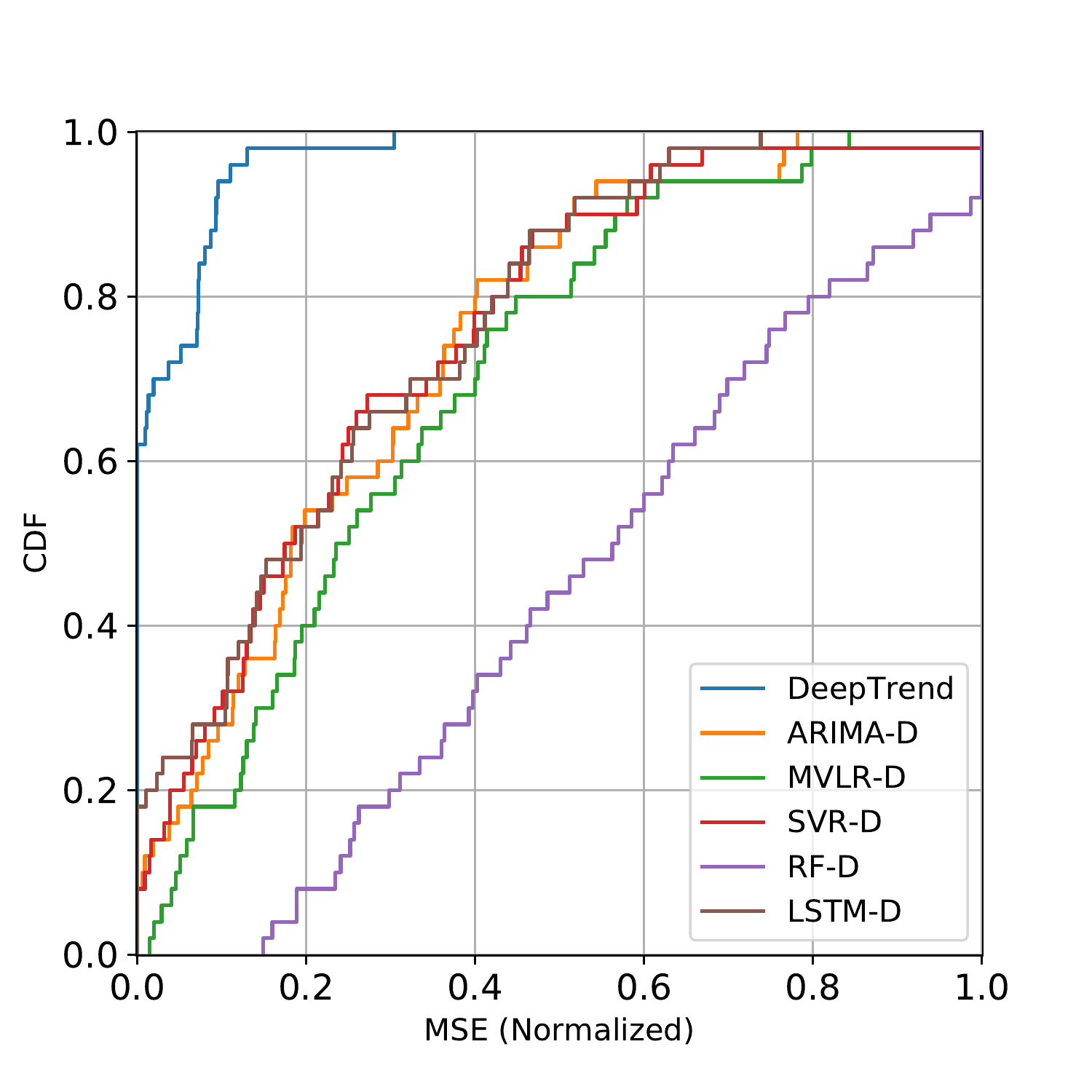}
	\caption{Empirical CDF of the MSE for 50 test stations.}
	\label{fig: CDF_MSE}
\end{figure}

\begin{figure}[!htb]
	\centering
	\includegraphics[width=3.5in]{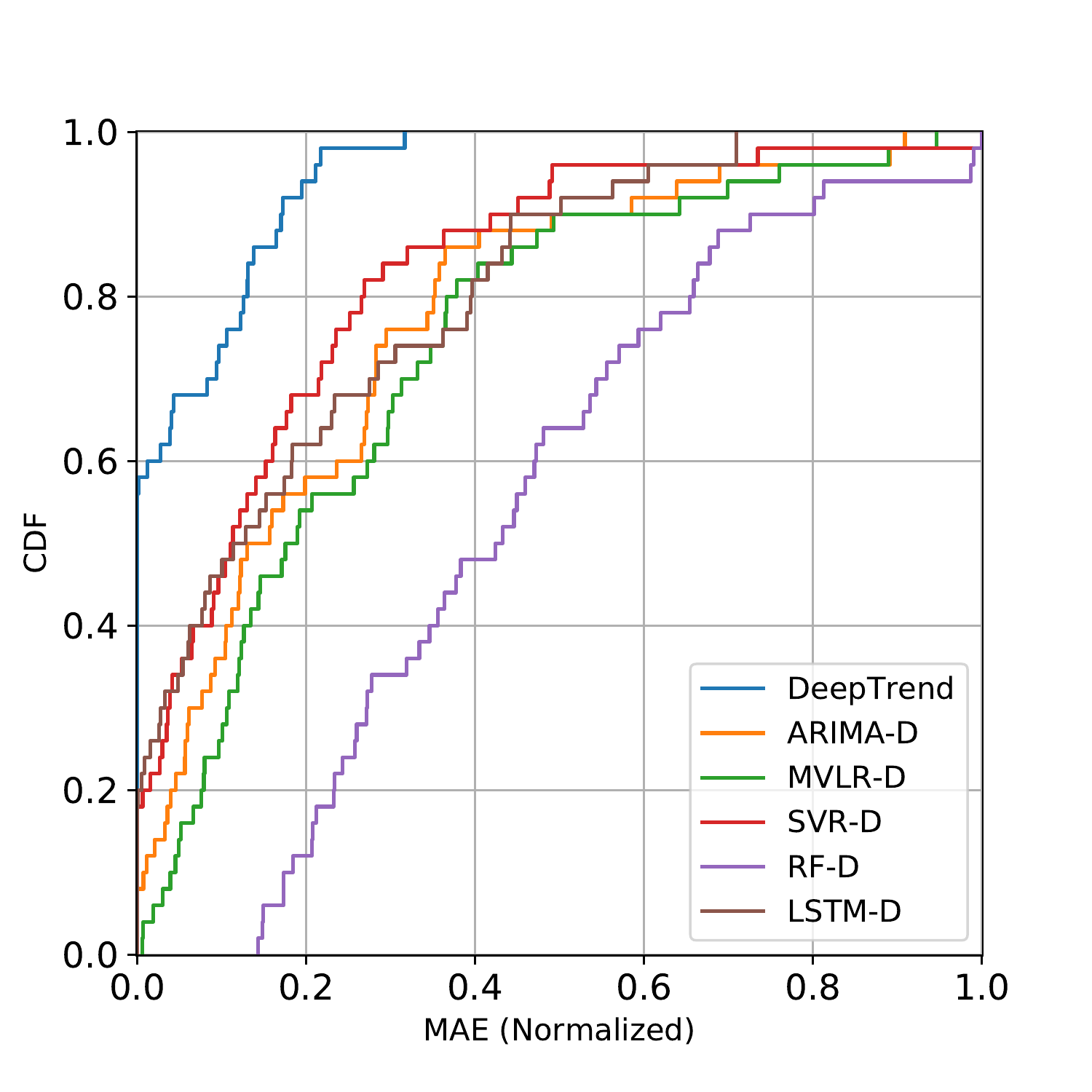}
	\caption{Empirical CDF of the MAE for 50 test stations.}
	\label{fig: CDF_MAE}
\end{figure}

As shown in Table \ref{tab: model_compare}, the proposed DeepTrend makes MSE and MAE drop to 984.47 and 21.21, respectively, which noticeably outperforms other models. A visual display of performance comparison is given in Fig. \ref{fig: CDF_MSE} and Fig. \ref{fig: CDF_MAE}. They present the cumulative distribution function (CDF) of MSE and MAE for DeepTrend and the detrending based models, which  describe the statistical results on 50 test stations. In figures, MSE and MAE have been first normalized between 0 and 1 for the test results of all models in each test station. We can find that DeepTrend outperform other models in terms of MSE and MAE in most of test stations, which demonstrates that the proposed model is effective and promising.

\section{Conclusion}
In this paper, we explore whether the deep network LSTM can learn the temporal pattern of traffic flow in prediction. Experiments reveal that simply using LSTM is not superior to some traditional machine learning models like SVR if detrending is not used, showing that it does not learn the patterns in traffic flow. To better capture the temporal pattern, we propose DeepTrend,  a deep hierarchical neural network which integrates the process of pattern extraction and flow prediction. Compared with traditional LSTM, DeepTrend needs pre-training layer-by-layer and then fine-tuning in the entire network. The first kind of layer extraction layer is used to learn the temporal pattern of flow series, and the second kind of layer prediction layer is to make a prediction for incoming flow which is fed by output series from extraction layer and calculated residual series. The experiments show that DeepTrend outperforms LSTM and other baselines based on detrending methods.

We only take account of temporal pattern in this paper. For future work, it would be considered that making the deep network learn the spatial correlations between different stations and integrating the temporal-spatial dependence in one network for traffic flow prediction.


%
%




%

%
%

\let\oldbibliography\thebibliography
\renewcommand{\thebibliography}[1]{\oldbibliography{#1}
\setlength{\itemsep}{10pt}} 

\bibliographystyle{IEEEtran}
\bibliography{Traffic-Prediction}

\end{document}